%% file: main.tex
\documentclass[runningheads]{llncs}
\usepackage[T1]{fontenc}
\usepackage{graphicx}
\usepackage{booktabs}
\usepackage[misc]{ifsym}

\usepackage{amsmath,amssymb,amsfonts}
\usepackage{cleveref}

\usepackage{xcolor}
\usepackage[linesnumbered,ruled,vlined,algo2e,noend]{algorithm2e}
\usepackage{amsmath}
\usepackage{algorithm}
\usepackage[noend]{algpseudocode}
\crefname{algocf}{alg.}{algs.}
\Crefname{algocf}{Algorithm}{Algorithms}
\usepackage{bbm}
\usepackage{multirow}
\usepackage{url}
\usepackage{tabularx}
\usepackage{microtype}

\SetCommentSty{mycommfont}
\SetKwInput{KwInput}{Input}                %
\SetKwInput{KwOutput}{Output}              %
\usepackage{algpseudocode}

\usepackage{listings}
\begin{document}

\title{MASCOTS: Model-Agnostic Symbolic COunterfactual explanations for Time Series}

\titlerunning{MASCOTS: Model-Agnostic Symbolic COunterfactual explanations}

\author{Dawid Płudowski\inst{1}\orcidID{0009-0001-2386-7497} \and \\
Francesco Spinnato\inst{3,4}\orcidID{0000-0002-3203-6716} \and \\
Piotr Wilczyński\inst{1,7}\orcidID{0009-0009-8418-3062} \and \\
Krzysztof Kotowski\inst{5}\orcidID{0000-0003-2596-6517} \and \\ 
Evridiki Vasileia Ntagiou\inst{6}\orcidID{0000-0003-3403-2863} \and \\
Riccardo Guidotti\inst{3,4}\orcidID{0000-0002-2827-7613} \and \\
Przemysław Biecek\inst{1, 2}\orcidID{0000-0001-8423-1823}}

\authorrunning{D. Pludowski et al.}

\institute{Warsaw University of Technology, Poland 
\and
University of Warsaw, Poland
\and
University of Pisa, Italy
\and
ISTI-CNR, Pisa, Italy
\and
 KP Labs, Poland
\and
 European Space Operations Centre, Germany
\and
ETH Zürich, Switzerland
}

\maketitle

\input{abstract}
\input{introduction}

\input{related_works}

\input{background}

\input{methodology}

\input{experiments}

\input{results}
\input{discussion}

\input{credits}

\input{refs.bbl}

\end{document}

%% file: abstract.tex
\begin{abstract}
Counterfactual explanations provide an intuitive way to understand model decisions by identifying minimal changes required to alter an outcome. 
However, applying counterfactual methods to time series models remains challenging due to temporal dependencies, high dimensionality, and the lack of an intuitive human-interpretable representation. 
We introduce MASCOTS, a method that leverages the Bag-of-Receptive-Fields representation alongside symbolic transformations inspired by Symbolic Aggregate Approximation. 
By operating in a symbolic feature space, it enhances interpretability while preserving fidelity to the original data and model.
Unlike existing approaches that either depend on model structure
or autoencoder-based sampling,
MASCOTS directly generates meaningful and diverse counterfactual observations in a model-agnostic manner, operating on both univariate and multivariate data. 
We evaluate MASCOTS on univariate and multivariate benchmark datasets, demonstrating comparable validity, proximity, and plausibility to state-of-the-art methods, while significantly improving interpretability and sparsity. 
Its symbolic nature allows for explanations that can be expressed visually, in natural language, or through semantic representations, making counterfactual reasoning more accessible and actionable.

\end{abstract}

%% file: introduction.tex
\section{Introduction}
Time series classification (TSC) plays a crucial role in various fields, including healthcare, climate science, and engineering.
Its wide-ranging applications have driven the development of increasingly powerful predictors capable of achieving remarkable classification accuracy. 
Both univariate and multivariate time series classification have gained significant research interest, as demonstrated by recent ``bake-offs''~\cite{ruiz2021great,middlehurst2024bake}, which periodically benchmark the top-performing classifiers. 
The findings from these evaluations are consistent: the most effective classifiers are powerful hybrid ensembles, such as MultiRocket-Hydra~\cite{dempster2023hydra}, Hive-Cote 2~\cite{middlehurst2021hive}, and InceptionTime~\cite{ismail2020inceptiontime}. 
These models leverage both prediction and feature spaces from multiple underlying algorithms, resulting in state-of-the-art accuracy. 
However, their complexity renders them black-box models, meaning they lack interpretability from a human perspective.

\begin{figure}[t]
    \centering
    \includegraphics[width=\linewidth]{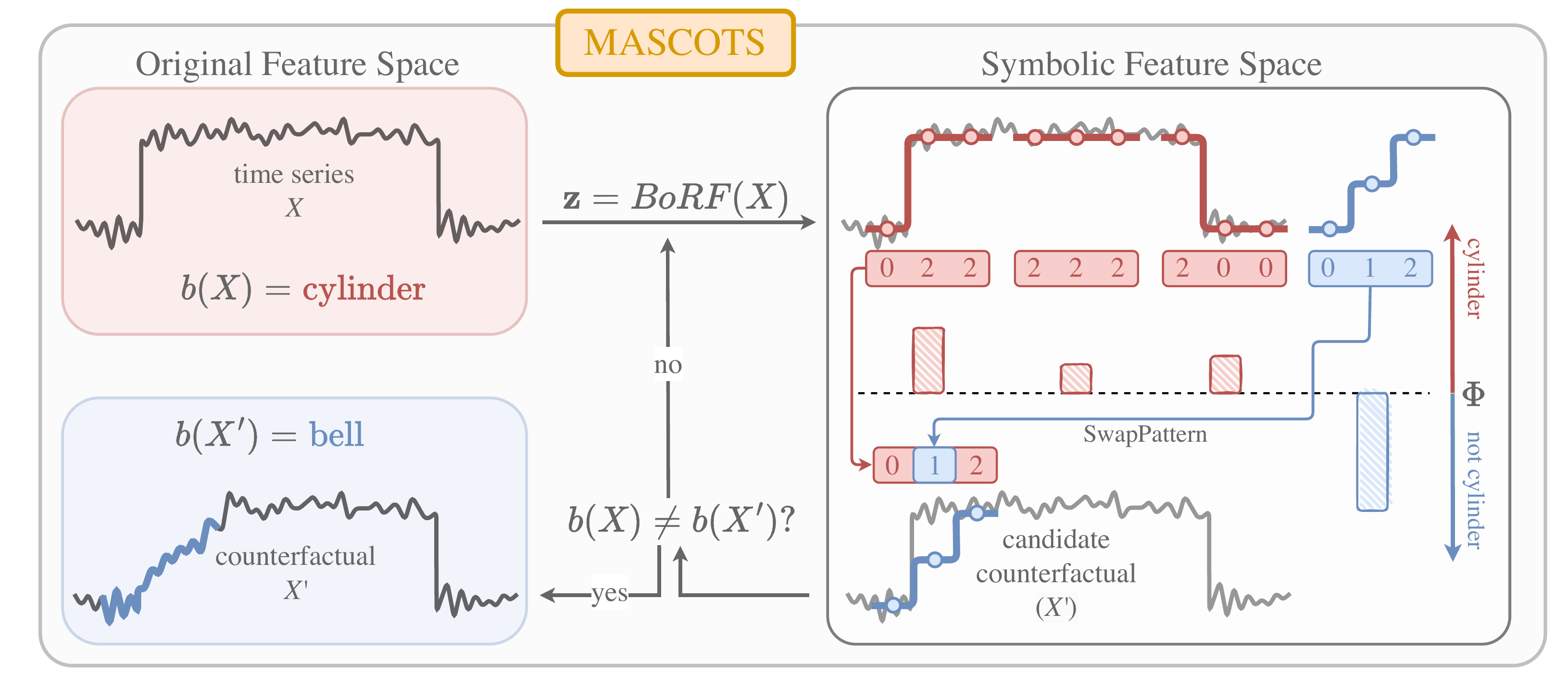}
    \caption{A simplified graphical abstract of \textsc{mascots}. Given a time series $X$ classified by a black-box model $b$ as \textit{cylinder}, $X$ is transformed into a symbolic representation, $\mathbf{z}$, using the Bag-of-Receptive-Fields (BoRF)~\cite{spinnato2024fast}. In this semantic space, $X$ is represented as symbolic subsequences. To identify a candidate counterfactual, the most important positive and negative patterns (\textit{0,2,2} and \textit{0,1,2} in the illustration) are selected. The negative pattern is then \textit{swapped} within the time series to generate a candidate counterfactual. If the black-box model's predicted class remains unchanged, the process repeats. Otherwise, the final counterfactual is returned.
    }
    \label{fig:abstract}
\end{figure}

This work aims to enhance the interpretability of black-box models in time series classification through Explainable Artificial Intelligence (XAI) techniques~\cite{bodria2023benchmarking}. 
While XAI offers diverse tools for explaining complex models, most approaches have been developed for tabular or image data, with time series explainability only recently gaining traction~\cite{theissler2022explainable}. 
In this domain, explanations commonly take the form of saliency maps~\cite{lundberg2017unified}, highlighting the most relevant part of observations contributing to the classification outcome, or subsequence-based explanations such as shapelets~\cite{ye2009time} focusing on significant sub-patterns within a time series. 
Instead, this work centers on instance-based explanations, where entire time series serve as the primary explanatory objects. 
Specifically, we focus on counterfactual explanations, i.e., minimal modifications to an input time series that alter the classification outcome of a black-box model~\cite{wachter2017counterfactual}. 
Counterfactual explanations are particularly powerful because, unlike feature importance methods~\cite{lundberg2017unified}, they indicate what must change in a given instance to achieve a different classification result~\cite{guidotti2024counterfactual}. They are useful as they facilitate reasoning about the cause-effect relationships between observed features and classification outputs. %

In this work, we introduce \textsc{mascots}, a novel model-agnostic method for generating counterfactual explanations in time series. \textsc{mascots} leverages symbolic representations to significantly enhance interpretability while maintaining fidelity. As depicted in \Cref{fig:abstract}, \textsc{mascots} initially converts a given time series into a Bag-of-Receptive-Fields (BoRF) representation~\cite{spinnato2024fast}, capturing essential symbolic patterns. It then identifies those patterns most strongly influencing classification outcomes, both positively and negatively. By iteratively substituting negative patterns, \textsc{mascots} modifies the original time series until it obtains a valid counterfactual, ensuring an interpretable and robust explanatory process.

Our contributions are as follows. \emph{(i)} Unlike existing methods that rely on model-specific architectures or autoencoder-based sampling~\cite{wang2021learning,spinnato2023understanding}, \textsc{mascots} operates in a fully transparent manner, making it broadly applicable to predictive time series models without additional constraints. 
\emph{(ii)} To the best of our knowledge, \textsc{mascots} is the first approach that employs a symbolic subsequence-based semantic space for explainability in the time series domain, providing a semantic-rich counterfactual generation process that does not depend on nearest unlike neighbors (NUN)~\cite{dasarathy1995nearest}. 
\emph{(iii)} \textsc{mascots} facilitates both visual and natural language counterfactual explanations, improving the interpretability of the counterfactual generation process. 
\emph{(iv)} Furthermore, \textsc{mascots} enables explanations for state-of-the-art ensemble models, a task where most existing methods fail due to their reliance on internal model structures. 
Through extensive evaluation on benchmark datasets, we demonstrate that \textsc{mascots} achieves comparable performance to state-of-the-art techniques regarding validity, proximity, and plausibility while significantly improving the sparsity of counterfactual explanations. 
By bridging symbolic representations with counterfactual reasoning, \textsc{mascots} represents a significant step forward in explainability for time series models. %

The structure of this work is as follows: \Cref{sec:related} examines related research on counterfactuals, while \Cref{sec:background} outlines the background of our proposed methodology. \Cref{sec:method} elaborates on the approach, followed by experimental results and analysis in \Cref{sec:experiments}. 
Lastly, \Cref{sec:conclusion} presents the conclusions.

%% file: related_works.tex
\section{Related Works}
\label{sec:related}
The simplest counterfactual models for time series rely on classical distance-based algorithms, such as K-Nearest Neighbors (KNN) or Nearest Unlike Neighbor (NUN)~\cite{dasarathy1995nearest}, which identify the closest existing instance of a different class. 
In this category, we find approaches like {Native Guides}~\cite{delaney2021instancebased} and TimeX~\cite{filali2022mining}, where time series are perturbed to generate counterfactuals while adhering to desirable properties such as proximity, sparsity, plausibility, and diversity. 
However, these approaches are limited to univariate data. 
CoMTE~\cite{ates2021counterfactual} extends this framework to multivariate data by modifying time series from the training set, and computing the minimal number of substitutions necessary to change the original classification. 
Since CoMTE primarily operates by identifying and substituting similar patterns from different classes, it may struggle to produce meaningful counterfactuals when the dataset lacks sufficiently close examples.
AB-CF~\cite{li2023attention} and DiscoX~\cite{bahri2024discord} take a different approach by focusing on local patterns. 
They extract fixed-length subsequences using a sliding window or identify discords via the matrix profile~\cite{yeh2016matrix}, replacing them with the nearest counterparts from the desired class. 
Given their emphasis on locality, these methods may fail to capture global temporal dependencies, potentially leading to counterfactuals that are locally valid but globally unrealistic. 
Finally, in~\cite{karlsson2020locally}, KNN is employed to locally and globally \emph{tweak} time series, altering the outcome of specific black-box models. However, this approach is also restricted to univariate data. Contrary to classical distance-based approaches, our proposal, \textsc{mascots}, does not rely on NUNs or Euclidean distance, as it performs perturbations in a semantic space produced by an interpretable transformation, i.e., the Bag-of-Receptive-Fields~\cite{spinnato2024fast}.

More complex approaches leverage evolutionary algorithms~\cite{hollig2022tsevo} and generative models, such as autoencoders. 
Autoencoders come in various forms, including recurrent neural networks, such as LSTMs, which have been used in~\cite{labaien2020contrastive} to extend the concept of Contrastive Explanation Methods to time series. %
Convolutional neural networks (CNNs) have also been utilized, as seen in LASTS~\cite{spinnato2023understanding}, as well as methods that test both, such as LatentCF++~\cite{wang2021learning}.
The central idea of these approaches is to perturb time series within a simplified latent space, ensuring that counterfactuals remain closer to the distribution of the training set. 
In this sense, autoencoders can play an indirect role as a loss component that assesses counterfactual \textit{plausibility}. 
Sub-SpaCE~\cite{refoyo2024sub} exemplifies this approach by evaluating the plausibility of counterfactuals generated through a genetic algorithm with tailored mutation and initialization strategies. 
This method encourages modifications in a minimal number of subsequences, producing highly sparse explanations. 
Plausibility is assessed similarly in TeRCE~\cite{bahri2022temporal}, which leverages the shapelet transform to identify the most relevant shapelets for a given class, pinpoint their locations in the input instance, and replace them with values derived from the NUN. %
While these methods are model-agnostic, their main limitation is the requirement to train a separate autoencoder for each dataset~\cite{spinnato2023understanding}, which makes their usage across a wide range of tasks impractical. 
In contrast, \textsc{mascots}, does not require any generative model to produce a counterfactual.

Counterfactual explanations can also be model-specific, i.e., designed for a particular black-box model. 
With the exception of~\cite{karlsson2020locally}, which targets the Random Shapelet Forest~\cite{karlsson2016generalized}, most model-specific approaches focus on explaining neural network-based methods. 
One such example is Glacier~\cite{wang2024glacier}, an extension of LatentCF++ designed to generate counterfactuals for any deep-learning-based model. 
While this approach is both promising and extensively tested, it is limited to univariate data. 
Other notable methods include CELS~\cite{li2023cels} for univariate data and M-CELS~\cite{li2024m} for multivariate data, both of which employ a gradient-based strategy. These methods utilize three interdependent modules to produce sparse counterfactuals, leveraging a learned saliency map to guide perturbations. 
The primary limitation of gradient-based model-specific approaches is that, while certain deep learning architectures are highly effective for TSC, the current state-of-the-art consists primarily of hybrid ensemble methods~\cite{middlehurst2024bake}. 
These ensembles often integrate multiple classifiers, making it difficult, or even impossible, to compute gradients, thus restricting the applicability of such techniques. \textsc{mascots} does not share these limitations as it is model-agnostic.

Finally, to the best of our knowledge, most of the aforementioned methods rely exclusively on visualizations to convey counterfactual modifications. 
In contrast, we enable the generation of counterfactual explanations in a structured, human-understandable manner using natural language descriptions. %
Specifically, given a time series $X$ and its counterfactual $X'$, the transformation can be articulated through a structured statement such as: \textit{``To change the prediction of a black-box model from class $c_i$ to $c_j$, the time series $X$ needs to contain pattern $a$ instead of $b$.''}
One approach that might seem similar is PUPAE~\cite{der2024pupae}, but the key difference is that it relies on a predefined set of templates that require domain expertise to construct. 
In contrast, \textsc{mascots} derives its explanations directly from the time series patterns, eliminating the need for manually designed templates. 
This makes \textsc{mascots} not only more flexible but also broadly applicable across different domains without requiring specialized knowledge.

%% file: background.tex
\section{Background}
\label{sec:background}

This section provides all the necessary concepts to understand our proposal. %

\begin{definition}[Time Series Data]
    A \textit{time series dataset}, $\mathcal{X}=\{X_{1}, \dots, X_{n}\} \in \mathbb{R}^{n \times d \times m}$, is a collection of $n$ time series. A time series, $X$, is a collection of $d$ signals (or channels), $X = \{\mathbf{x}_{1}, \dots, \mathbf{x}_{d}\} \in {\mathbb{R}}^{d \times m}$. A signal, $\mathbf{x}$, is a sequence of $m$ real-valued observations sampled regularly,
    $\mathbf{x} = [x_{1}, \dots, x_{m}] \in {\mathbb{R}}^{m}$.
\end{definition}
\noindent
When $d=1$, the time series is \textit{univariate}, for $d > 1$ it is \textit{multivariate}. 
Time series datasets can be used in a variety of tasks. 
This work focuses on supervised learning, particularly Time Series Classification (TSC).

\begin{definition}[Time Series Classification]
Let $\mathcal{X}$ be a time series dataset and $\mathbf{y} \in \{1, \dots, c\}^n$ its corresponding labels vector, where $c$ is the number of classes. The goal of \textit{Time Series Classification} is to train a model $f$ that maps each time series $X_i \in \mathcal{X}$ to a predicted label $\hat{y}_i$, such that $f(X_i) = \hat{y}_i$ for all $i \in \{1, \dots, n\}$. This yields the predicted label vector $\mathbf{\hat{y}} = [\hat{y}_1, \dots, \hat{y}_n] \in \{1, \dots, c\}^{n}$.
\end{definition}

\noindent The goal of time series classification is to ensure that the trained model $f$ predicts a label $\hat{\mathbf{y}}$ that closely matches the true labels $\mathbf{y}$, typically by minimizing a classification loss function during training. Many models produce probability distributions over classes, i.e., $\hat{Y} \in [0,1]^{n \times c}$, with $\hat{\mathbf{y}}$ determined by the highest probability. %
While maximizing accuracy is important, providing explanations for the predictions of a given model is becoming more and more relevant.
This work focuses on a specific type of explanation, i.e., \emph{counterfactual explanations}. 
Counterfactual time series show the minimal changes in the input data that lead to a different decision outcome~\cite{theissler2022explainable}. 
Formally:
\begin{definition}[Counterfactual]
    Given a classifier $f$ that outputs the decision $\hat{y} = f(X)$ for an instance $X$, a \emph{counterfactual} consists of an instance $X'$ such that the decision for $f$ on $X'$ is different from $\hat{y}$, i.e., $f(X') \neq \hat{y}$, and such that $X'$ is \emph{similar} to $X$, and that $X'$ is \emph{plausible}.
\end{definition}
Similarity (or proximity) usually refers to a distance metric, while plausibility depends on the counterfactual domain and is assessed by verifying that the instance is not merely an adversarial example and remains semantically coherent with the dataset. 
Other relevant metrics include \textit{sparsity}, i.e., the number of features altered to generate a counterfactual, and \textit{validity}, i.e., the ability of a counterfactual method to produce a valid counterfactual. 

Counterfactuals can be obtained in several ways; here, we propose to adopt \emph{surrogate models}.
Explanations based on surrogate methods can clarify the behavior of black-box models, $b$, by employing a secondary, more interpretable model, $g$, to approximate the behavior of the primary black-box model, i.e., $b(X) \simeq g(X)$~\cite{bodria2023benchmarking}. 
By doing so, the surrogate model seeks to provide insights into the complex model’s decisions by mimicking its outputs while remaining inherently more interpretable. 
A great advantage of surrogates is that they are model-agnostic, i.e., they can explain any black-box without any assumption about its inner components. Surrogates can be trained on the original raw time series data or after processing it into a more interpretable tabular representation.

In this work, we adopt the Bag-of-Receptive-Fields (BoRF)~\cite{spinnato2024fast} as our interpretable tabular representation. BoRF extends the classical Bag-of-Patterns~\cite{baydogan2013bag} and, akin to the Bag-of-Words approach in text analysis, converts a time series into a vector of pattern counts. This transformation is achieved by sliding a potentially strided and dilated window along the time series to extract all possible receptive fields of a specified length, $w$.
In this context, a receptive field is just a time series subsequence that can have gaps inside, i.e., can skip observations. 
These subsequences are then standardized and discretized into \emph{words} of length $l\leq w$ using the Symbolic Aggregate Approximation (SAX)~\cite{lin2007experiencing}. 
SAX uses the Piecewise Aggregate Approximation (PAA)~\cite{keogh2001dimensionality} to segment each subsequence into equal-sized segments and then compute the mean value for each segment. 
Finally, these values are quantized using a set of breakpoints, obtained through the quantiles of the standard Gaussian distribution, which bin values in equiprobable symbols, $\alpha$.
Thus, from each time series, $X$, a set of patterns is extracted, where a single pattern is denoted as $\mathbf{p}=[\alpha_1,\dots,\alpha_l]\in\mathbb{A}^l$, with $\mathbb{A}$ being a set of finite symbols. 
Each symbolic pattern is bidirectionally hashed into an integer $k$, allowing for both encoding and decoding, and is then stored in the Bag-of-Receptive-Fields. 
Formally:
\begin{definition}[Bag-of-Receptive-Fields]
\label{def:borf}
    Given a time series dataset $\mathcal{X} \in {\mathbb{R}}^{n \times d \times m}$ and a set of $h$ patterns, a Bag-of-Receptive-Fields is a tensor $\mathcal{Z}\in\mathbb{N}^{n\times d \times h}$, where $z_{i,j,k}$ is the number of appearances of the hashed SAX pattern $k$ in the signal $j$ of time series $i$.
\end{definition}

\noindent The Bag-of-Receptive-Fields $\mathcal{Z}$ can be flattened into $Z\in\mathbb{N}^{n\times r}$, where $r$ is the total number of patterns across channels, i.e., $r=dh$. An example of such a representation is shown in \Cref{fig:abstract} (top-right), where a time series is represented as four patterns (counts are omitted for better readability). $Z$ can then be used as a training set for any standard tabular classifier, $g$, offering the advantage of interpretable features, specifically, the count of occurrences of each pattern within a time series. 
Finally, the classifier can be interpreted using any standard explainer, such as feature importance-based methods like SHAP~\cite{lundberg2017unified}.

\begin{definition}[Feature Importance]
    Given a single row of $Z$, $\mathbf{z}\in\mathbb{R}^{r}$, a \textit{feature importance} matrix, $\Phi = [\boldsymbol\phi_1, \dots, \boldsymbol\phi_{r}]\in\mathbb{R}^{r\times c}$, contains the contribution of each feature value $z\in\mathbf{z}$ towards predicting each possible class $c$.
\end{definition}

\noindent In the following section, we exploit feature importance in the Bag-of-Receptive-Fields semantic space, and propose a counterfactual technique based on symbolic patterns, which iteratively produces interpretable perturbations to generate counterfactual explanations for time series black-box classifiers.

%% file: methodology.tex
\section{Methodology}
\label{sec:method}

In this section, we introduce \textsc{mascots}, a \textsc{m}odel-\textsc{a}gnostic \textsc{s}ymbolic \textsc{co}unterfactual explanation method for \textsc{t}ime \textsc{s}eries classification. \textsc{mascots} combines the Bag-of-Receptive-Fields (BoRF) representation with symbolic transformations inspired by Symbolic Aggregate Approximation (SAX) to enhance interpretability while maintaining fidelity to the original data and model.

\textsc{mascots} takes as input a time series $X$, a surrogate model $g$, a training dataset $\mathcal{X}$ for the surrogate, a black-box model $b$, an attribution method $e$, and a penalty hyperparameter $\lambda$. 
In essence, \textsc{mascots} trains a surrogate model on a Bag-of-Receptive-Fields representation of the time series dataset. 
This surrogate serves as an interpretable proxy for the black-box, allowing the extraction of pattern relevance for classification using a feature attribution method. The resulting feature importance then guides semantic perturbations of the time series by modifying symbolic words. The approach is detailed step-by-step in the following sections, and illustrated in \Cref{fig:abstract}.

\subsection{Counterfactual Generation Process}
The pseudo-code of \textsc{mascots} is reported in \Cref{alg:cf}.
The first step is to train an interpretable surrogate model (lines 1-3). 
To achieve this, \textsc{mascots} requires a training set $\mathcal{X}$, and obtains the corresponding black-box predictions, $\hat{\mathbf{y}}=b(\mathcal{X})$ (line 1). 
Next, the training data $\mathcal{X}$ is transformed into a Bag-of-Receptive-Fields representation, $Z$, using BoRF~\cite{spinnato2024fast} (line 2). 
Finally, the surrogate model is trained on $Z$ and $\hat{\mathbf{y}}$, effectively learning to approximate the black-box predictions on the given dataset (line 3). 
The black-box is also used to predict the label of the time series whose prediction we are explaining, i.e., $\hat{y}=b(X)$ (line 4). 
Then, $X'$, a copy of $X$, is created (line 5), and the counterfactual generation loop begins.  
The condition of the counterfactual generation loop (line 6) checks whether the black-box prediction on the perturbed time series $X'$ matches that of the original time series $X$.
While this is true, the loop proceeds as follows. $X'$ is converted into a Bag-of-Receptive-Fields vector, $\mathbf{z}$, (line 7), the attribution method is used to produce a feature importance matrix, $\Phi$, containing the contribution of each pattern value in $\mathbf{z}$ towards each class (line 8). 
Then, the GetPerturbation procedure is invoked to generate the perturbation (line 9), which is subsequently added to the time series (line 10). 
After this, the loop condition is re-evaluated, and the final counterfactual, $X'$, is eventually returned.

\begin{figure}[t]
    \centering
    \includegraphics[width=0.9\linewidth]{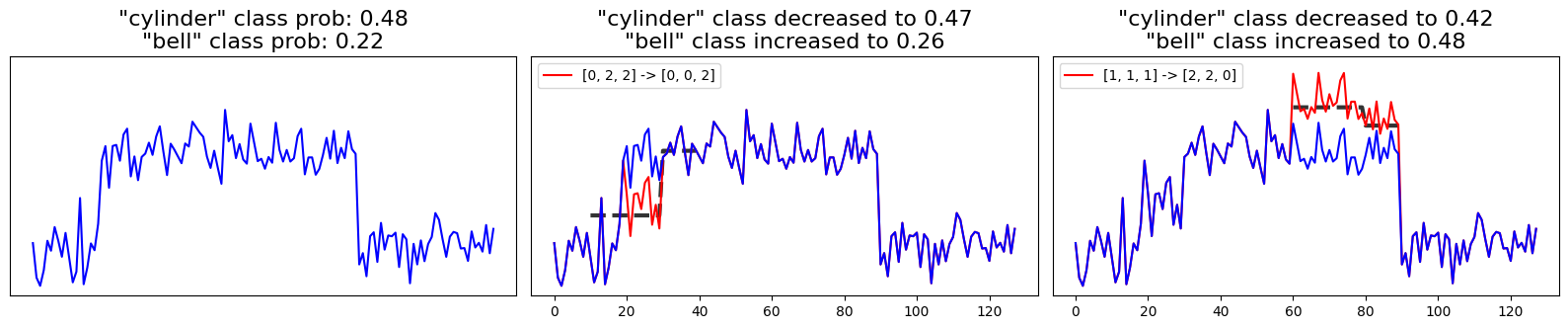}
    \caption{Example of the iterative process of \textsc{mascots}, where a time series is incrementally perturbed from a \textit{cylinder} to a \textit{bell}. Inserted patterns are marked by the black dashed line.}
    \label{fig:example}
\end{figure}

Using an iterative algorithm reinforces the \textbf{validity} of generated counterfactuals by gradually shifting the black-box model's prediction toward the target class, as illustrated in \Cref{fig:example}. However, multiple changes, even if meaningful, may overly distort the counterfactual, reducing plausibility. Similar to~\cite{wang2024glacier}, we set a task-dependent iteration limit to balance these factors, halting the algorithm regardless of success. As shown in \Cref{sec:experiments}, the required iterations remain low on average, preventing excessive modifications.

\begin{algorithm2e}[t]
    \caption{\textsc{mascots}}\label{alg:cf}
    
    \KwData{$X$ - time series to explain, 
    $\mathcal{X}$ - surrogate training dataset, 
    $g$ - surrogate model, 
    $b$ - black-box model, 
    $e$ - attribution method, 
    $\lambda$ - penalty}
    \KwResult{Counterfactual $X'$}
    
    $\mathbf{\hat{y}} \gets b(\mathcal{X})$; \tcp*[f]{\scriptsize Predict classes of surrogate dataset}\\
    $Z \gets BoRF(\mathcal{X});$\tcp*[f]{\scriptsize Convert surrogate dataset into Bag-of-Receptive-Fields}\\
    $g \gets train(g, Z, \mathbf{\hat{y}})$;\tcp*[f]{\scriptsize Train surrogate}\\
    $\hat{y} \gets b(X)$; \tcp*[f]{\scriptsize Predict class of time series to explain}\\
    $X' \gets X$\;
    \While(\tcp*[f]{\scriptsize While the black-box prediction does not change}){$b(X') = \hat{y}$}{
        $\mathbf{z} \gets BoRF(X')$;\tcp*[f]{\scriptsize Convert time series into Bag-of-Receptive-Fields}\\
        $\Phi \gets e(g, \mathbf{z})$;\tcp*[f]{\scriptsize Get attribution matrix}\\
        $\Delta \gets \text{GetPerturbation}(X', \hat{y}, \mathbf{z}, \Phi, \lambda)$;\tcp*[f]{\scriptsize Generate perturbation}\\
        $X' \gets X'+\Delta$;\tcp*[f]{\scriptsize Perturb time series}\\
    }
    \Return{$X'$}
\end{algorithm2e}

\subsection{Pattern Swapping}
\label{sec:methodology-swap}
We illustrate the GetPerturbation procedure in \Cref{alg:swap}.
It begins by utilizing the feature attribution matrix to identify the most important pattern for the predicted class, denoted as $k^+$. 
This corresponds to the symbolic word that has the highest relevance toward the classification outcome assigned to the time series $X$ by the black-box model (line 1). 
We restrict the search to \textit{contained patterns}, i.e., patterns where $z_k>0$, as we want to perturb an existing pattern in $X$.
Using the inverse hash function provided by BoRF, $k^+$ is then mapped back to its corresponding pattern vector (line 2), $\mathbf{p}^+$. 
Next, GetPerturbation identifies the most relevant pattern index opposing the predicted class, $k^-$, i.e., the index of the symbolic word that most strongly influences a classification different from $\hat{y}$ (line 3). 
Similar to line 2, the corresponding pattern vector, $\mathbf{p}^-$, is retrieved (line 4).
The swapping step alters only these two patterns while the others remain unchanged, thus intrinsically encouraging the perturbation's \textbf{sparsity}.
The penalty parameter $\lambda$ is applied to penalize patterns that deviate significantly from $\mathbf{p}^+$ to ensure that the transformation remains meaningful and does not introduce too unrealistic modifications. 
A high value of $\lambda$ further encourages selecting similar patterns, reducing the number of altered elements in the time series. This constraint enhances the \textbf{proximity} property of the counterfactual, ensuring that modifications remain minimal.
Proximity and sparsity, paired with the fact that the swap is performed between two patterns that both exist in the training dataset,  $\mathcal{X}$, push the generated perturbation to remain within a reasonable semantic range, i.e., they promote \textbf{plausibility}.  
The vector $\mathbf{p}^+$ is then aligned to the time series to determine the channel, 
$j$, and starting timestamp, $t$, in $X'$ where the perturbation will be applied (line 5). 
Since a given pattern can have multiple valid alignments within the time series, a random index is selected from the available options to introduce variability while maintaining realism. 
Finally, the perturbation matrix, $\Delta$, is initialized (line 6) and populated using the PatternSwap function (line 7). 

\begin{algorithm2e}[t]
    \caption{GetPerturbation}\label{alg:swap}
    \KwData{$X$ - time series to swap, 
    $\hat{y}$ - predicted class, 
    $\mathbf{z}$ Bag-of-Receptive-Fields, 
    $\Phi$ - attribution map, 
    $\lambda$ - penalty}
    \KwResult{Perturbation - $\Delta$}
    $k^+ \gets \arg\max\limits_{k:z_k\neq0}\phi_{k,\hat{y}}$; \tcp*[f]{\scriptsize Id of most important word for prediction}\\
    $\mathbf{p}^+ \gets hash^{-1}(k^+)$; \tcp*[f]{\scriptsize Retrieve important word for prediction}\\
    $k^- \gets \arg\min\limits_k\phi_{k,\hat{y}}+\lambda \|\mathbf{p}^+ - \mathbf{p}_k\|_1$; \tcp*[f]{\scriptsize Id of most important word against prediction}\\
    $\mathbf{p}^- \gets hash^{-1}(k^-)$; \tcp*[f]{\scriptsize Retrieve most important word against prediction}\\
    $j,t \gets align(X,\mathbf{p}^+)$; \tcp*[f]{\scriptsize Find pattern channel and timestamp alignment}\\
    $\Delta\gets\mathbf{0}^{d\times m}$; \tcp*[f]{\scriptsize Initialize empty perturbation matrix}\\
    $\Delta_{j,t:t+w} \gets \text{PatternSwap}(X, \Delta,\mathbf{p}^+,\mathbf{p}^-)$; \tcp*[f]{\scriptsize Swap pattern}\\
    \Return{$\Delta$}
\end{algorithm2e}

The PatternSwap function operates on a time series $X$ and takes as input the two SAX patterns, $\mathbf{p^+}=[\alpha^+_1,\dots,\alpha^+_l]$ and $\mathbf{p^-}=[\alpha^-_1,\dots,\alpha^-_l]$. 
Its primary objective is to perturb the subsequence from channel $j$, starting at index $t$, $\mathbf{x}_{j,t:t+w}=[x_{j,t},\dots,x_{j,t+w}]$, which corresponds to the pattern $\mathbf{p^+}$, so that when SAX is applied, this subsequence is instead encoded as $\mathbf{p^-}$. 
Let $\mu$ and $\sigma$ be the mean and standard deviation of the subsequence, and let $\bar{\mathbf{x}}=[\bar{x}_1,\dots,\bar{x}_l]$ be its segmented representation obtained via PAA, where each segment consists of $w/l$ observations, which are averaged within that segment.
To achieve the transformation, we first define the perturbation needed to shift the segmented value $\bar{x}_i$ from its original symbolic representation $\alpha^+_i$ to the target $\alpha^-_i$. 
This perturbation is given by the difference between $\bar{x}_i$ and the central value of the bin corresponding to $\alpha^-_i$, denoted as $q^{\alpha^-_i}$. 
The perturbation is then denormalized using the inverse standardization formula, incorporating the subsequence's mean $\mu$ and standard deviation $\sigma$. 
Formally, %
\begin{equation}
    \delta_i = (q^{\alpha^-_i} - \bar{x}_i) \sigma + \mu,
\end{equation}

\noindent where $\delta_i$ represents changes that must be added to all observations corresponding to $\bar{x}_i$ to move them into a different breakpoint bin. 
To apply this perturbation to the original subsequence $\mathbf{x}_{j,t:t+w}$, we produce $\boldsymbol{\delta}=[\delta_1,\dots,\delta_1,\dots,\delta_l,\dots,\delta_l]$ such that each $\delta_i$ is repeated $w/l$ times. %
Thus, the perturbation is simply $\Delta_{j,t:t+w} = \boldsymbol{\delta}$, which can be summed to the original time series $X$ (line 10, \Cref{alg:cf}). %
This transformation ensures that the local structure of the time series is preserved while allowing flexible modifications. 
Importantly, the locality of perturbations is not strictly enforced. A single perturbation can potentially influence an extended portion of the time series, depending on the subsequence length corresponding to the most important pattern $\mathbf{p}^+$. Even if the final modification to the time series is relatively large, it remains interpretable, as it can be succinctly described using only $l$ values, where each subsequence segment of length $w/l$ is shifted by a single scalar, reducing cognitive complexity.

%% file: experiments.tex
\section{Experiments}
\label{sec:experiments}

In this section, we assess \textsc{mascots} on both univariate and multivariate time series classification datasets from the UEA and UCR repositories~\cite{dau2019ucr}, comparing its performance against state-of-the-art methods. We use the datasets listed in~\Cref{tab:data}, all of which have been featured in multiple studies introducing counterfactuals for time series~\cite{wang2024glacier,li2024m}. %

As baseline counterfactual explainers, we employ Glacier~\cite{wang2024glacier} and M-CELS~\cite{li2024m}, two recently proposed methods that achieve strong results without requiring extensive training of generative adversarial networks or multiple runs of genetic algorithms. For Glacier, we use its ``uniform'' variant, which offers the best trade-off in terms of \textit{proximity} and \textit{sparsity}. Gradients are computed in the latent space of a 1D-CNN autoencoder, as this setup has been shown to yield the highest validity scores. Due to Glacier’s limitations, we restrict its evaluation to univariate data.
For M-CELS, we adhere to the hyperparameter settings recommended by the authors, except for disabling \texttt{tvnorm} and \texttt{budget}, which we empirically found to improve validity. %
We use InceptionTime~\cite{ismail2020inceptiontime} as a black-box model due to its state-of-the-art performance in TSC and ability to provide model gradients required by M-CELS and Glacier. While \textsc{mascots} does not rely on gradients, selecting InceptionTime ensures a fair comparison. Additionally, we include a qualitative example with MultiRocket-Hydra~\cite{dempster2023hydra}, a model that lacks gradient access and is therefore incompatible with most counterfactual explainers.

\textsc{mascots} is parametrized by the choice of $\lambda$, the surrogate model, the attribution method, and the configuration of the BoRF transformation. In our experiments, we set $\lambda \in \{0.0, 0.1\}$. The surrogate model is a shallow neural network with two layers, capable of training on probabilities rather than discrete labels. For attribution, we use SHAP~\cite{lundberg2017unified}.
BoRF automatically tailors its main hyperparameters for each dataset, accounting for the number of points and channels. To maintain subsequence contiguity, dilation is fixed at $1$, the alphabet size is set to $3$, and the stride is chosen as $w/l$ (the ratio of word size to word length) for efficient transformation. Other parameters, including the number of SAX configurations, word size $w$, and length $l$, are dynamically adjusted to capture both local and global patterns. For each dataset, the smallest SAX words contain at least $8$ points, while the largest are the highest power of two not exceeding the time series length. Each SAX word consists of either $2$ or $4$ symbols. The algorithm is limited to a maximum of $20$ iterations to mitigate adversarial effects.

\begin{table}[t]
\begin{center}
\caption{Datasets description follows the notation introduced in~\Cref{sec:background}: $n$ (instances), $d$ (channels), $m$ (points), and $c$ (classes). $ACC(b)$ and $FID(g)$ denote the accuracy of InceptionTime and the fidelity of \textsc{mascots}' surrogate, respectively.}
\label{tab:data}
\begin{tabularx}{10.6cm}{@{}p{3cm} >{\raggedleft\arraybackslash}p{1cm} >{\raggedleft\arraybackslash}p{1cm} >{\raggedleft\arraybackslash}p{1cm} >{\raggedleft\arraybackslash}p{1cm} >{\raggedleft\arraybackslash}p{1.5cm} >{\raggedleft\arraybackslash}p{1.5cm}@{}}
\toprule
Dataset                 & $n$ & $d$ & $m$ & $c$ & $ACC(b)$ & $FID(g)$ \\ \midrule
TwoLeadECG                & $23$            & $1$           & $82$            & $2$          &  $1.00$            & $1.00 $             \\
GunPoint                  & $50$           & $1$           & $150$           & $2$          & $1.00$          & $1.00$              \\
Earthquakes               & $322$          & $1$           & $512$           & $2$          & $0.68$           & $0.82$              \\
Coffee                    & $28$           & $1$           & $286$           & $2$          & $1.00$           & $1.00$              \\
Wine                      & $57$           & $1$           & $234$           & $2$          & $0.80$           & $0.84$              \\
ItalyPowerDemand          & $67$           & $1$           & $24$            & $2$          & $0.97$           & $0.91$              \\
BasicMotions              & $40$           & $6$           & $100$           & $4$          & $0.50$           & $1.00$              \\
Cricket                   & $108$          & $6$           & $1197$          & $12$         & $0.24$               & $0.60$                  \\
Epilepsy                  & $137$          & $3$           & $206$           & $4$          & $0.30$           & $0.90$              \\
RacketSports              & $151$          & $6$           & $30$            & $4$          & $0.37$           & $0.64$              \\ \bottomrule
\end{tabularx}
\end{center}
\end{table}

\paragraph{\textbf{Evaluation Measures.}}

We adopt the primary measures of counterfactual quality reported in prior studies~\cite{wang2024glacier,li2023cels}: \textit{validity}, \textit{proximity}, \textit{sparsity}, and \textit{plausibility}. The first three are formally defined below, while plausibility is assessed using Isolation Forest~\cite{liu2008isolation}, where we report the fraction of counterfactuals classified as nominal (non-outliers). Additionally, we provide runtime comparisons for all algorithms using the same hardware setup\footnote{System: 8 cores AMD Rome 7742, 32GB RAM.}.  

Validity measures the proportion of generated counterfactuals $X_i'$ that lead to a different classifier prediction compared to the original instance $X_i$. It is defined as $ \mathit{validity}(\mathcal{X}, \mathcal{X}') = \frac{1}{n}\sum_{i=1}^{n} \mathbbm{1}[f(X_i) \neq f(X_i')] $, where $\mathbbm{1}[f(X_i) \neq f(X_i')]$ is an indicator function that equals $1$ if the model's prediction changes and $0$ otherwise. A high validity score indicates that most counterfactuals effectively alter the model’s decision.  
Proximity quantifies the average distance between original instances and their corresponding counterfactuals: $ \mathit{proximity}(\mathcal{X}, \mathcal{X}') = \frac{1}{n \cdot d \cdot m} \sum^n_{i=1}\| X_i - X_i' \| $, where $\| X_i - X_i' \|$ represents the distance between $X_i$ and $X_i'$. Lower proximity values indicate that counterfactuals remain close to the original data points, making them more realistic and interpretable.  
Sparsity captures the fraction of features that remain unchanged between the original and counterfactual instances: $ \mathit{sparsity}(\mathcal{X}, \mathcal{X}') = \frac{1}{n \cdot d \cdot m} \sum^n_{i=1}\sum^d_{j=1} \sum^m_{t=1}  \mathbbm{1}[ x_{i,j,t} - x_{i,j,t}'  = 0] $, where $\mathbbm{1}[ x_{i,j,t} - x_{i,j,t}'  = 0]$ equals $1$ if a feature remains unchanged and $0$ otherwise. Higher sparsity values indicate that fewer features are modified, promoting more interpretable counterfactual explanations.

\begin{figure}[t]
    \centering
    \includegraphics[trim={0 20mm 0 0},clip,width=\linewidth]{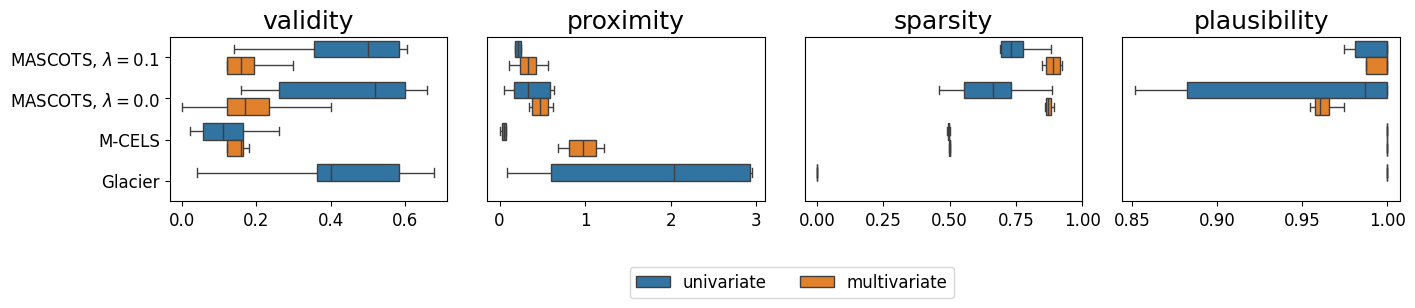}
    \includegraphics[trim={0 5mm 0 70mm},clip,width=\linewidth]{results-comp-metrics-boxplot.png}
    \caption{Box-plots of evaluation measures. For \textit{validity}, \textit{sparsity}, and \textit{plausibility}, the higher score is better, while for \textit{proximity}, the smaller the better. 
    \textsc{mascots} stands out in \textit{sparsity} with a decent \textit{validity} and \textit{proximity}. The difference between \textsc{mascots}-$\lambda=0.0$ and \textsc{mascots}-$\lambda=0.1$ suggests a trade-off between validity and other measures.}
    \label{fig:results-boxplots}
\end{figure}

\begin{table}[t]
\begin{center}
\caption{Mean and standard deviation for the various evaluation measures aggregated over univariate and multivariate datasets.
For \textsc{mascots}, we add the average number of iterations required to flip a label. Best values are highlighted in bold. 
}
\label{tab:results}
\setlength{\tabcolsep}{1.1mm}
\begin{tabular}{ccccccc} %
\toprule
                                     &  & \textit{validity}$\uparrow$    & \textit{proximity} $\downarrow$ & \textit{sparsity} $\uparrow$  & \textit{plausibility} $\uparrow$ & \# iter. $\downarrow$ \\ \midrule
\multirow{4}{*}{\rotatebox{90}{univariate}} & \textsc{mascots}$_{\lambda=0.1}$        & $\mathbf{0.44 \pm 0.18}$                          & $0.24 \pm 0.18$                         & $\mathbf{0.71 \pm 0.12}$                        & $0.98 \pm 0.02$                        & $3.0 \pm 1.0$    \\
                                     & \textsc{mascots}$_{\lambda=0.0}$        & $\mathbf{0.44 \pm 0.21}$                          & $0.35 \pm 0.25$                         & $0.65 \pm 0.15$                        & $0.84 \pm 0.29$    & $\mathbf{2.7 \pm 1.3}$                        \\
                                     & M-CELS                        & $0.11 \pm 0.08$                          & $\mathbf{0.08 \pm 0.10}$                         & $0.49 \pm 0.00$                        & $0.99 \pm 0.01$      &  $-$                    \\
                                     & Glacier                       & $0.41 \pm 0.23$                          & $2.32 \pm 2.34$                         & $0.00 \pm 0.00$                        & $\mathbf{1.00 \pm 0.00}$        & $-$                     \\ \midrule
\multirow{3}{*}{\rotatebox{90}{multiv.}}                & \textsc{mascots}$_{\lambda=0.1}$        & $0.15 \pm 0.12$                          & $\mathbf{0.33 \pm 0.19}$                         & $\mathbf{0.88 \pm 0.03}$                        & $0.98 \pm 0.02$                     & $\mathbf{1.7 \pm 2.7}$       \\
                                     & \textsc{mascots}$_{\lambda=0.0}$        & $\mathbf{0.18 \pm 0.16}$                          & $0.47 \pm 0.13$                         & $0.87 \pm 0.01$                        & $0.96 \pm 0.00$                         & $2.4 \pm 3.6$   \\
                                     & M-CELS                        & $0.12 \pm 0.08$                          & $0.96 \pm 0.23$                         & $0.49 \pm 0.00$                        & $\mathbf{1.00 \pm 0.00}$                           &  $-$ \\ \bottomrule
\end{tabular}
\end{center}
\end{table}

\paragraph{\textbf{Results.}}

The experimental results for \textsc{mascots} and the baseline models are illustrated through box plots in \Cref{fig:results-boxplots} and summarized with mean and standard deviation values in \Cref{tab:results}.  
For \textit{validity}, \textsc{mascots} and Glacier perform particularly well on univariate data, achieving approximately $41\%$–$44\%$ valid counterfactuals on average. On multivariate datasets, both configurations of \textsc{mascots} outperform M-CELS. Notably, setting the $\lambda$ parameter to $0.0$ slightly improves the validity of counterfactuals generated by \textsc{mascots}.  
In terms of \textit{proximity}, \textsc{mascots} and M-CELS produce counterfactuals that remain reasonably close to the original observations, while Glacier generates counterfactuals that are significantly more distant. For multivariate datasets, \textsc{mascots} consistently outperforms M-CELS in proximity.  
Regarding \textit{sparsity}, \textsc{mascots} excels, altering fewer than $30\%$ of the original features on average. In contrast, Glacier, due to its gradient-based nature, modifies every point at least slightly, preventing it from generating sparse explanations. M-CELS, on the other hand, alters approximately $50\%$ of the time series to construct a counterfactual.  
Furthermore, we report the average number of iterations (i.e., the number of pattern swaps) required by \textsc{mascots} to generate a counterfactual. This number varies across datasets and configurations but typically does not exceed $3$. This suggests that, on average, \textsc{mascots} can produce effective counterfactuals with only three meaningful semantic modifications. 
Finally, regarding runtime, performance is comparable with an average of 23 minutes for \textsc{mascots}, 26 minutes for Glacier, and 55 minutes for M-CELS.

\begin{figure}[t]
    \centering
    \includegraphics[width=\linewidth]{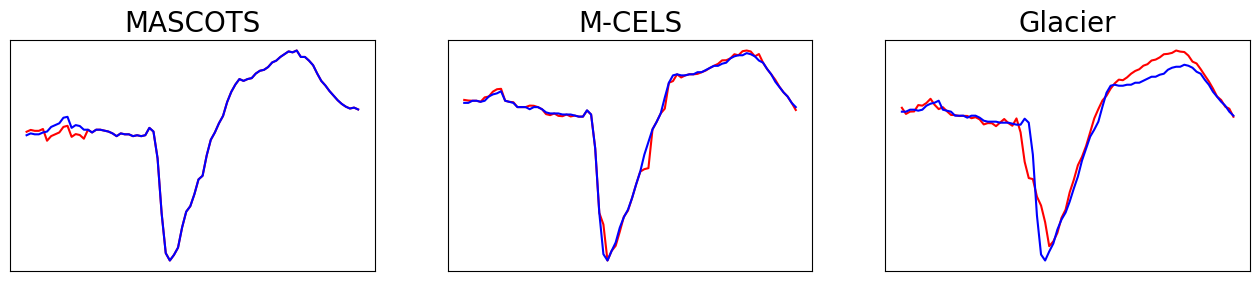}
    \caption{Example of \textsc{mascots} on the \textit{TwoLeadECG} dataset to explain InceptionTime. 
    \textsc{mascots} is able to create sparse counterfactual which maintain its local structure. On the other hand, M-CELS and Glacier produce small (perhaps adversarial) undesirable changes to the original time series, varying in this way its initial shape.}
    \label{fig:results-comp}
\end{figure}

\paragraph{\textbf{Qualitative Examples.}}
In \Cref{fig:results-comp}, we present the counterfactuals (in red) generated by each analyzed method for a randomly selected time series (in blue) from the \textit{TwoLeadECG} dataset~\cite{dau2019ucr}, which represents heart ECG signals.
For \textsc{mascots}, we set $\lambda=0.1$ to achieve better proximity and sparsity. 
In this case, \textsc{mascots} generates a counterfactual by modifying only a single pattern at the beginning of the signal while preserving its local structure.
The changes produced by M-CELS are also minimal but are distributed across the signal. 
In contrast, Glacier introduces significant alterations to the original observation, disrupting the local structure of the signal.

\begin{figure}[t]
    \centering
    \includegraphics[width=\linewidth]{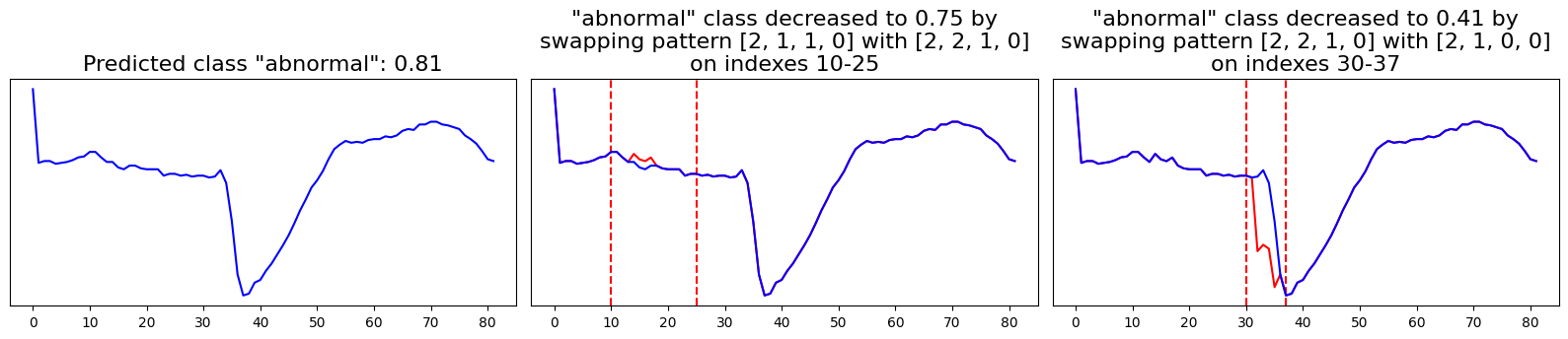}
    \caption{Example of \textsc{mascots} on the \textit{TwoLeadECG} dataset to explain MultiRocket-Hydra. Changes are presented both as visualization and natural language. 
    If positive and negative patterns have corresponding symbols (for example, the 1st, 3rd, and 4th symbols in the middle plot), \textsc{mascots} does not change them.}
    \label{fig:results-hydra}
\end{figure}

We also provide an example on \textit{TwoLeadECG}, focusing on explaining the MultiRocket-Hydra black-box model~\cite{dempster2023hydra}. Unlike fully neural network-based models, MultiRocket-Hydra incorporates non-neural components, making methods such as Glacier and M-CELS inapplicable.  
The explanation, provided both visually and in natural language, is illustrated in \Cref{fig:results-hydra}. The counterfactual is generated in two iterations. First, \textsc{mascots} introduces a small modification between indexes $10$ and $25$. While this initial change does not significantly affect the black-box model’s prediction, it alters the feature importance matrix $\mathit{\Phi}$, enabling \textsc{mascots} to identify a more impactful transformation in the second iteration. This final modification broadens the valley in the signal, ultimately flipping the model’s classification from “abnormal” to “normal.”  
The generated counterfactual can be expressed in natural language as follows: \textit{``To obtain a counterfactual for an abnormal ECG, the pattern in indexes $10$–$25$ must be replaced with $[2,2,1,0]$, followed by replacing the pattern in indexes $30$–$37$ with $[2,1,0,0]$.''} These patterns differ in length—$16$ and $8$, respectively—demonstrating the flexibility of \textsc{mascots} in adapting its transformations.  
In both this example and \Cref{fig:results-hydra}, the symbols $0, 1, 2$ correspond to ``low,'' ``medium,'' and ``high'' values in the time series, respectively, providing an intuitive interpretation of the modifications.

%% file: discussion.tex
\section{Conclusion}
\label{sec:conclusion}

In this article, we have introduced \textsc{mascots}, a model-agnostic method for generating counterfactual explanations in univariate and multivariate time series classification. 
By leveraging the BoRF transformation and symbolic representations, \textsc{mascots} enhances interpretability while maintaining fidelity to the original black-box model. 
Unlike prior approaches, it operates in a fully agnostic manner without the need of autoencoders and without relying on distances or nearest unlike neighbors. 
Our evaluation demonstrates its effectiveness, achieving high interpretability and sparsity while preserving validity and proximity.

For future work, we aim to extend \textsc{mascots} into a ``user-in-the-middle'' framework, allowing expert intervention at each iteration to refine counterfactual explanations. 
By selecting among proposed changes, experts can enhance the plausibility of generated counterfactuals, explore custom ``what-if?'' scenarios, and even create artificial observations for manual labeling and integration into training datasets. 
This interactive approach could further improve the adaptability and utility of counterfactual explanations in real-world applications.

%% file: credits.tex
\begin{credits}

\subsubsection{\discintname}
The authors have no competing interests to declare that are relevant to the content of this article.
\end{credits}

%% file: main.bbl
\begin{thebibliography}{10}
\providecommand{\url}[1]{\texttt{#1}}
\providecommand{\urlprefix}{URL }
\providecommand{\doi}[1]{https://doi.org/#1}

\bibitem{ates2021counterfactual}
Ates, E., Aksar, B., Leung, V.J., Coskun, A.K.: Counterfactual explanations for multivariate time series. In: 2021 international conference on applied artificial intelligence (ICAPAI). pp.~1--8 (2021)

\bibitem{bahri2022temporal}
Bahri, O., Li, P., Boubrahimi, S.F., Hamdi, S.M.: Temporal rule-based counterfactual explanations for multivariate time series. In: 2022 21st IEEE International Conference on Machine Learning and Applications (ICMLA). pp. 1244--1249 (2022)

\bibitem{bahri2024discord}
Bahri, O., Li, P., Filali~Boubrahimi, S., Hamdi, S.M.: Discord-based counterfactual explanations for time series classification. Data Mining and Knowledge Discovery  \textbf{38}(6),  3347--3371 (2024)

\bibitem{baydogan2013bag}
Baydogan, M.G., Runger, G., Tuv, E.: A bag-of-features framework to classify time series. IEEE transactions on pattern analysis and machine intelligence  \textbf{35}(11),  2796--2802 (2013)

\bibitem{bodria2023benchmarking}
Bodria, F., Giannotti, F., Guidotti, R., Naretto, F., Pedreschi, D., Rinzivillo, S.: Benchmarking and survey of explanation methods for black box models. Data Mining and Knowledge Discovery pp. 1--60 (2023)

\bibitem{dasarathy1995nearest}
Dasarathy, B.V.: Nearest unlike neighbor (nun): an aid to decision confidence estimation. Optical Engineering  \textbf{34}(9),  2785--2792 (1995)

\bibitem{dau2019ucr}
Dau, H.A., Bagnall, A., Kamgar, K., Yeh, C.C.M., Zhu, Y., Gharghabi, S., Ratanamahatana, C.A., Keogh, E.: The ucr time series archive. IEEE/CAA Journal of Automatica Sinica  \textbf{6}(6),  1293--1305 (2019)

\bibitem{delaney2021instancebased}
Delaney, E., Greene, D., Keane, M.T.: Instance-based counterfactual explanations for time series classification. In: Case-Based Reasoning Research and Development. pp. 32--47 (2021)

\bibitem{dempster2023hydra}
Dempster, A., Schmidt, D.F., Webb, G.I.: Hydra: Competing convolutional kernels for fast and accurate time series classification. Data Mining and Knowledge Discovery  \textbf{37}(5),  1779--1805 (2023)

\bibitem{der2024pupae}
Der, A., Yeh, C.C.M., Zheng, Y., Wang, J., Zhuang, Z., Wang, L., Zhang, W., Keogh, E.: Pupae: Intuitive and actionable explanations for time series anomalies. In: Proceedings of the 2024 SIAM International Conference on Data Mining (SDM). pp. 37--45 (2024)

\bibitem{filali2022mining}
Filali~Boubrahimi, S., Hamdi, S.M.: On the mining of time series data counterfactual explanations using barycenters. In: Proceedings of the 31st ACM International Conference on Information \& Knowledge Management. pp. 3943--3947 (2022)

\bibitem{guidotti2024counterfactual}
Guidotti, R.: Counterfactual explanations and how to find them: literature review and benchmarking. Data Mining and Knowledge Discovery  \textbf{38}(5),  2770--2824 (2024)

\bibitem{hollig2022tsevo}
H{\"o}llig, J., Kulbach, C., Thoma, S.: Tsevo: Evolutionary counterfactual explanations for time series classification. In: 2022 21st IEEE International Conference on Machine Learning and Applications (ICMLA). pp. 29--36 (2022)

\bibitem{ismail2020inceptiontime}
Ismail~Fawaz, H., Lucas, B., Forestier, G., Pelletier, C., Schmidt, D.F., Weber, J., Webb, G.I., Idoumghar, L., Muller, P.A., Petitjean, F.: Inceptiontime: Finding alexnet for time series classification. Data Mining and Knowledge Discovery  \textbf{34}(6),  1936--1962 (2020)

\bibitem{karlsson2016generalized}
Karlsson, I., Papapetrou, P., Bostr{\"o}m, H.: Generalized random shapelet forests. Data mining and knowledge discovery  \textbf{30},  1053--1085 (2016)

\bibitem{karlsson2020locally}
Karlsson, I., Rebane, J., Papapetrou, P., Gionis, A.: Locally and globally explainable time series tweaking. Knowl. Inf. Syst.  \textbf{62}(5),  1671--1700 (2020)

\bibitem{keogh2001dimensionality}
Keogh, E., Chakrabarti, K., Pazzani, M., Mehrotra, S.: Dimensionality reduction for fast similarity search in large time series databases. Knowledge and information Systems  \textbf{3},  263--286 (2001)

\bibitem{labaien2020contrastive}
Labaien, J., Zugasti, E., De~Carlos, X.: Contrastive explanations for a deep learning model on time-series data. In: International Conference on Big Data Analytics and Knowledge Discovery. pp. 235--244 (2020)

\bibitem{li2023attention}
Li, P., Bahri, O., Boubrahimi, S.F., Hamdi, S.M.: Attention-based counterfactual explanation for multivariate time series. In: International Conference on Big Data Analytics and Knowledge Discovery. pp. 287--293 (2023)

\bibitem{li2023cels}
Li, P., Bahri, O., Boubrahimi, S.F., Hamdi, S.M.: Cels: Counterfactual explanations for time series data via learned saliency maps. In: 2023 IEEE International Conference on Big Data (BigData). pp. 718--727 (2023)

\bibitem{li2024m}
Li, P., Bahri, O., Boubrahimi, S.F., Hamdi, S.M.: M-cels: Counterfactual explanation for multivariate time series data guided by learned saliency maps. arXiv preprint arXiv:2411.02649  (2024)

\bibitem{lin2007experiencing}
Lin, J., Keogh, E., Wei, L., Lonardi, S.: Experiencing sax: a novel symbolic representation of time series. Data Mining and knowledge discovery  \textbf{15},  107--144 (2007)

\bibitem{liu2008isolation}
Liu, F.T., Ting, K.M., Zhou, Z.H.: Isolation forest. In: 2008 eighth ieee international conference on data mining. pp. 413--422 (2008)

\bibitem{lundberg2017unified}
Lundberg, S.M., Lee, S.I.: A unified approach to interpreting model predictions. Advances in neural information processing systems  \textbf{30} (2017)

\bibitem{middlehurst2021hive}
Middlehurst, M., Large, J., Flynn, M., Lines, J., Bostrom, A., Bagnall, A.: Hive-cote 2.0: a new meta ensemble for time series classification. Machine Learning  \textbf{110}(11),  3211--3243 (2021)

\bibitem{middlehurst2024bake}
Middlehurst, M., Sch{\"a}fer, P., Bagnall, A.: Bake off redux: a review and experimental evaluation of recent time series classification algorithms. Data Mining and Knowledge Discovery  \textbf{38}(4),  1958--2031 (2024)

\bibitem{refoyo2024sub}
Refoyo, M., Luengo, D.: Sub-space: Subsequence-based sparse counterfactual explanations for time series classification problems. In: World Conference on Explainable Artificial Intelligence. pp. 3--17 (2024)

\bibitem{ruiz2021great}
Ruiz, A.P., Flynn, M., Large, J., Middlehurst, M., Bagnall, A.: The great multivariate time series classification bake off: a review and experimental evaluation of recent algorithmic advances. Data mining and knowledge discovery  \textbf{35}(2),  401--449 (2021)

\bibitem{spinnato2024fast}
Spinnato, F., Guidotti, R., Monreale, A., Nanni, M.: Fast, interpretable, and deterministic time series classification with a bag-of-receptive-fields. {IEEE} Access  \textbf{12},  137893--137912 (2024)

\bibitem{spinnato2023understanding}
Spinnato, F., Guidotti, R., Monreale, A., Nanni, M., Pedreschi, D., Giannotti, F.: Understanding any time series classifier with a subsequence-based explainer. ACM Transactions on Knowledge Discovery from Data  \textbf{18}(2),  1--34 (2023)

\bibitem{theissler2022explainable}
Theissler, A., Spinnato, F., Schlegel, U., Guidotti, R.: Explainable ai for time series classification: A review, taxonomy and research directions. IEEE Access  (2022)

\bibitem{wachter2017counterfactual}
Wachter, S., Mittelstadt, B., Russell, C.: Counterfactual explanations without opening the black box: Automated decisions and the gdpr. Harv. JL \& Tech.  \textbf{31}, ~841 (2017)

\bibitem{wang2024glacier}
Wang, Z., Samsten, I., Miliou, I., Mochaourab, R., Papapetrou, P.: Glacier: guided locally constrained counterfactual explanations for time series classification. Machine Learning pp. 1--31 (2024)

\bibitem{wang2021learning}
Wang, Z., Samsten, I., Mochaourab, R., Papapetrou, P.: Learning time series counterfactuals via latent space representations. In: Discovery Science: 24th International Conference, DS 2021. pp. 369--384 (2021)

\bibitem{ye2009time}
Ye, L., Keogh, E.: Time series shapelets: a new primitive for data mining. In: Proceedings of the 15th ACM SIGKDD international conference on Knowledge discovery and data mining. pp. 947--956 (2009)

\bibitem{yeh2016matrix}
Yeh, C.C.M., Zhu, Y., Ulanova, L., Begum, N., Ding, Y., Dau, H.A., Silva, D.F., Mueen, A., Keogh, E.: Matrix profile i: all pairs similarity joins for time series: a unifying view that includes motifs, discords and shapelets. In: 2016 IEEE 16th international conference on data mining (ICDM). pp. 1317--1322 (2016)

\end{thebibliography}
